\documentclass[conference]{IEEEtran}
\IEEEoverridecommandlockouts
\usepackage{cite}
\usepackage{amsmath,amssymb,amsfonts}
\usepackage{algorithmic}
\usepackage{graphicx}
\usepackage{textcomp}
\usepackage{xcolor}
\usepackage{caption}
\usepackage{algorithm}
\usepackage{algorithmic}

\usepackage{booktabs}

\def\BibTeX{{\rm B\kern-.05em{\sc i\kern-.025em b}\kern-.08em
    T\kern-.1667em\lower.7ex\hbox{E}\kern-.125emX}}
\begin{document}

\title{From classical techniques to convolution-based models: A review of object detection algorithms

\thanks{The first author contributed the most to this paper. Corresponding author: mamiruzzaman@wcupa.edu}
}
\author{\IEEEauthorblockN{1\textsuperscript{st} FNU Neha}
\IEEEauthorblockA{\textit{Dept. of Computer Science} \\
\textit{Kent State University}\\
Kent, OH, USA\\
neha@kent.edu}
\and
\IEEEauthorblockN{1\textsuperscript{st} Deepshikha Bhati}
\IEEEauthorblockA{\textit{Dept. of Computer Science} \\
\textit{Kent State University}\\
Kent, OH, USA \\
dbhati@kent.edu}
\and
\IEEEauthorblockN{2\textsuperscript{nd} Deepak Kumar Shukla}
\IEEEauthorblockA{\textit{Rutgers Business School} \\
\textit{Rutgers University}\\
Newark, New Jersey, USA  \\
ds1640@scarletmail.rutgers.edu}
\and
\IEEEauthorblockN{3\textsuperscript{rd} Md Amiruzzaman}
\IEEEauthorblockA{\textit{Dept. of Computer Science} \\
\textit{West Chester University}\\
West Chester, PA, USA \\
mamiruzzaman@wcupa.edu}
}



\maketitle

\begin{abstract}
Object detection is a fundamental task in computer vision and image understanding, with the goal of identifying and localizing objects of interest within an image while assigning them corresponding class labels. Traditional methods, which relied on handcrafted features and shallow models, struggled with complex visual data and showed limited performance. These methods combined low-level features with contextual information and lacked the ability to capture high-level semantics. Deep learning, especially Convolutional Neural Networks (CNNs), addressed these limitations by automatically learning rich, hierarchical features directly from data. These features include both semantic and high-level representations essential for accurate object detection. This paper reviews object detection frameworks, starting with classical computer vision methods. We categorize object detection approaches into two groups: (1) classical computer vision techniques and (2) CNN-based detectors. We compare major CNN models, discussing their strengths and limitations. In conclusion, this review highlights the significant advancements in object detection through deep learning and identifies key areas for further research to improve performance.

\end{abstract}

\begin{IEEEkeywords}
Object Detection, CNN, Deep Learning, Image Processing, Computer Vision
\end{IEEEkeywords}

\section{INTRODUCTION}

\begin{figure}
    \centering
    \includegraphics[width=1\linewidth]{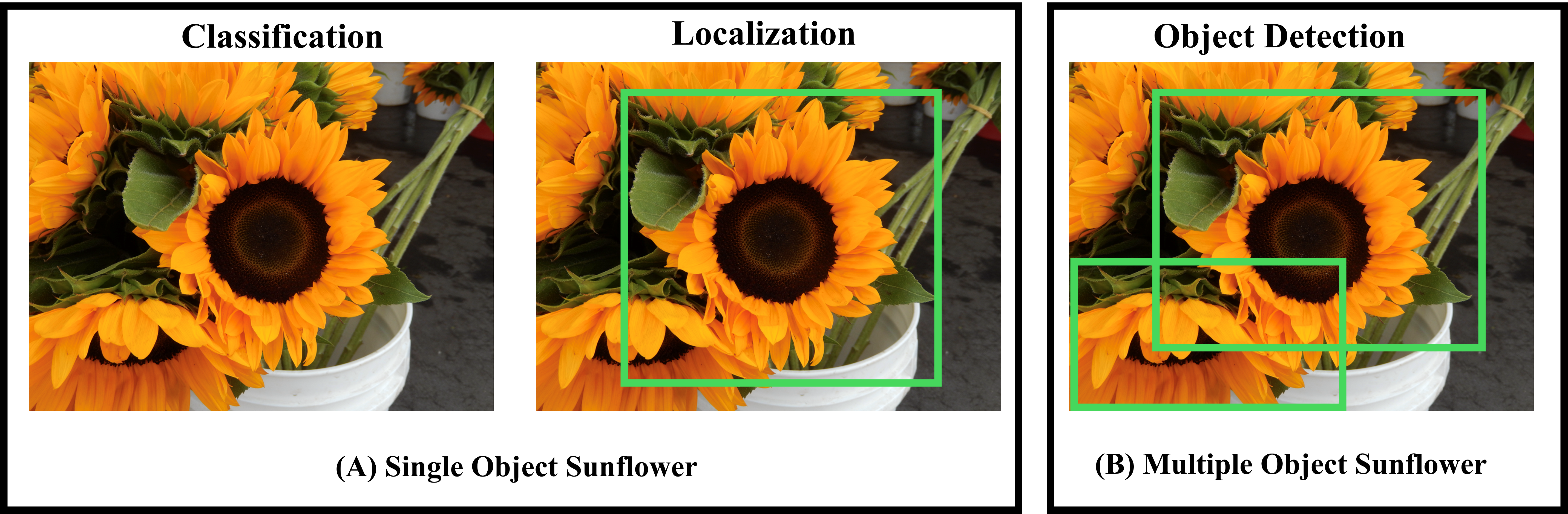}
    \caption{(A) Single-object sunflower: A single bounding box localizes and classifies the central sunflower bloom. (B) Multiple-object sunflower: Multiple bounding boxes highlight and classify overlapping sunflowers and leaves, illustrating multi-scale object detection and localization within a complex scene.}
    \label{fig:architecture0}
\end{figure}

Deep learning (DL) has advanced image analysis, especially in object classification, localization, and detection tasks. In classification, the aim is to assign an image or object within it to one of several categories \cite{chen2021review}. However, classification does not provide the object’s location. Localization improves on this by identifying both the object’s category and position, typically with a bounding box \cite{murthy2020investigations}, though the precision of these boxes can vary. Object detection further extends classification and localization by detecting and classifying multiple objects in an image, providing bounding boxes for each \cite{murthy2020investigations}. The bounding box’s top-left corner is represented by $(X_{min}, Y_{min})$, and the bottom-right by $(X_{max}, Y_{max})$, along with a label indicating the object’s class as shown in Fig \ref{fig:architecture0}.

Object detection has applications across fields such as medical imaging, logo detection, facial recognition, pedestrian detection, and industrial automation. However, challenges arise from image transformations like changes in scale, orientation, and lighting. While classical computer vision techniques provided a foundation, advancements in deep learning (DL), especially CNNs, have significantly improved detection performance. Modern methods use hierarchical representations, enabling object detection in complex environments with occlusions and varying scales.

Although many studies have reviewed specific deep learning models or object detection applications, few provide a comprehensive overview of both classical computer vision techniques and CNN-based approaches. This paper addresses this gap by offering an analysis of both. Key contributions include:

\begin{enumerate} 
    \item A review of classical computer vision techniques for object detection. 
    \item An analysis of general region proposal generation techniques. 
    \item A detailed review of convolution-based models for object detection, including two-stage and one-stage detectors. 
\end{enumerate}

The paper is organized as follows: Section 2 covers classical computer vision techniques for object detection, Section 3 discusses region proposal generation, Section 4 explores CNN-based detection architectures, Section 5 reviews applications, Section 6 lists popular datasets, Section 7 covers evaluation metrics, and Section 8 concludes with future directions.

\section{CLASSICAL COMPUTER VISION TECHNIQUES FOR OBJECT DETECTION}

Earlier computer vision techniques for image processing, particularly image similarity, relied on feature-based methods \cite{ma2021image, lowe2004distinctive, canny1986computational, dalal2005histograms, viola2001rapid, felzenszwalb2008discriminatively}. These methods focused on extracting distinctive image features to reduce computational costs while enabling robust image matching despite transformations like scaling or rotation \cite{ma2021image}. The Scale-Invariant Feature Transform (SIFT) algorithm overcame the challenge of scaling by extracting features invariant to scale, rotation, brightness, and contrast \cite{lowe2004distinctive}. Other feature extractors, like the Canny Edge Detector, contributed to tasks like image comparison and panoramic stitching by providing resilience to transformations and occlusions \cite{canny1986computational}. The Histogram of Oriented Gradients (HOG) technique enabled efficient image analysis by measuring gradient magnitudes and directions, creating descriptive feature vectors \cite{dalal2005histograms}.

Traditional object detection involves three stages:
\begin{enumerate}
    \item \textbf{Proposal Generation:} Scanning the image at various positions and scales to generate candidate bounding boxes, often using methods like sliding windows or selective search algorithms.
    \item \textbf{Feature Extraction:} Extracting features from the identified regions to capture relevant visual patterns.
    \item \textbf{Classification:} Classifying the extracted features using machine learning algorithms, such as support vector machine (SVM).
\end{enumerate}

In 2001, Viola et al. introduced a real-time (webcam based) facial detection classifier \cite{viola2001rapid}. In 2005, Dalal et al. introduced an object detector using HOG features and an SVM classifier, effective across scales but limited by pose variations \cite{dalal2005histograms}. In 2009, Felzenszwalb et al. improved this with the Deformable Part Model (DPM), allowing flexible parts to handle poses, though it struggled with overlapping parts in multi-person images \cite{felzenszwalb2008discriminatively}.

Studies from 2008 to 2012 on popular object detection datasets (see Section 5) showed key limitations in traditional methods. For instance, sliding windows require substantial computational resources and can generate redundant detections. Additionally, the performance of the classifier greatly impacts the results, necessitating more robust approaches.

\section{GENERIC REGION PROPOSAL GENERATION TECHNIQUES}

Object detection models integrate a bounding box regressor within the classification network to accurately locate objects \cite{schulter2014accurate}. Traditionally, this involves feeding cropped
images to the localization network, resulting in excessive inputs. An OverFeat model enhances efficiency by using a sliding window detector within convolution layers, scanning images with a large filter and stride \cite{sermanet2013overfeat}. However, indiscriminate scanning of background regions necessitates predicting potential object locations. Methods such as interest point detection, multiscale saliency, color contrast, edge detection, and super-pixel clustering are employed for this purpose \cite{li2015visual,fu2013salient,zitnick2014edge,fu2013superpixel}.

For instance, multiscale saliency leverages the Fast Fourier Transform to analyze features at multiple scales \cite{li2015visual}; color contrast relies on color intensity differences \cite{fu2013salient}; edge detection identifies edges, followed by density analysis \cite{zitnick2014edge}; and super-pixel clustering groups similar pixels for detailed analysis \cite{fu2013superpixel}.

Each method has specific limitations: multiscale saliency struggles with low-contrast objects, color contrast is ineffective with minimal contrast, edge detection may produce false positives or negatives, and super-pixel clustering requires refinement. Consequently, hybrid models are often developed to improve region proposal accuracy.

\section{CONVOLUTION BASED OBJECT DETECTION MODELS}

Object detection initially relied on manual feature design, focusing on patterns and edges. With CNN advancements, networks such as Visual Geometry Group Network (VGGNet) \cite{simonyan2014very} and AlexNet \cite{krizhevsky2012imagenet} now autonomously extract features through convolution and pooling layers, with fully connected (FC) layers followed by a SoftMax layer for classification. For localization, the final FC layer outputs bounding box coordinates, unlike classical methods which use filters and machine learning-based models (e.g., SVMs).

Training CNN-based models involve adjusting weights via backpropagation to align predictions with ground truth bounding boxes. Detection models fall into two categories: (1) Two-stage detectors, which generate region proposals before classification, including R-CNN \cite{girshick2014rich}, Fast R-CNN \cite{girshick2015fast}, Faster R-CNN \cite{ren2016faster}, and Mask R-CNN \cite{he2017mask}; and (2) One-stage detectors, treating detection as direct regression or classification tasks, like YOLO \cite{redmon2016you} and SSD \cite{liu2016ssd}. Table~\ref{table:cnn_comparison} summarizes their strengths and limitations.

\begin{table}[htbp]
\centering
\caption{Comparison of CNN-Based Object Detection Architectures}
\label{table:cnn_comparison}
\resizebox{\columnwidth}{!}{%
\begin{tabular}{|l|p{3cm}|p{3cm}|}
\hline
\textbf{Model} & \textbf{Strengths} & \textbf{Limitations} \\
\hline
R-CNN (2013) & 
Simple, foundational; applies CNNs for classification. & 
High computation for 2000 region classifications; slow (47 sec/image); no end-to-end training. \\
\hline
SPPNet (2015) & 
Faster than R-CNN; supports multi-scale input via spatial pyramid pooling. & 
Does not update conv. layers before SPP layer during fine-tuning. \\
\hline
Fast R-CNN (2015) & 
Faster than SPPNet; introduces ROI pooling to handle varied input sizes. & 
Relies on selective search for region proposals, not learned during training. \\
\hline
Faster R-CNN (2015) & 
Uses RPN for fast region proposals; improves efficiency. & 
Limited in detecting small objects due to single feature map. \\
\hline
Mask R-CNN (2017) & 
Adds instance segmentation, detecting objects and masks simultaneously. & 
High computational demand; struggles with motion blur at low resolution. \\
\hline
YOLO (2015) & 
Real-time detection at 45 fps; single forward pass. & 
Poor detection of small objects; produces coarse features. \\
\hline
SSD (2016) & 
Handles various resolutions; uses multi-scale feature maps for detection. & 
Default boxes may not match all shapes; possible overlapping detections. \\
\hline
\end{tabular}
}
\end{table}

\begin{figure}
    \centering
    \includegraphics[width=1\linewidth]{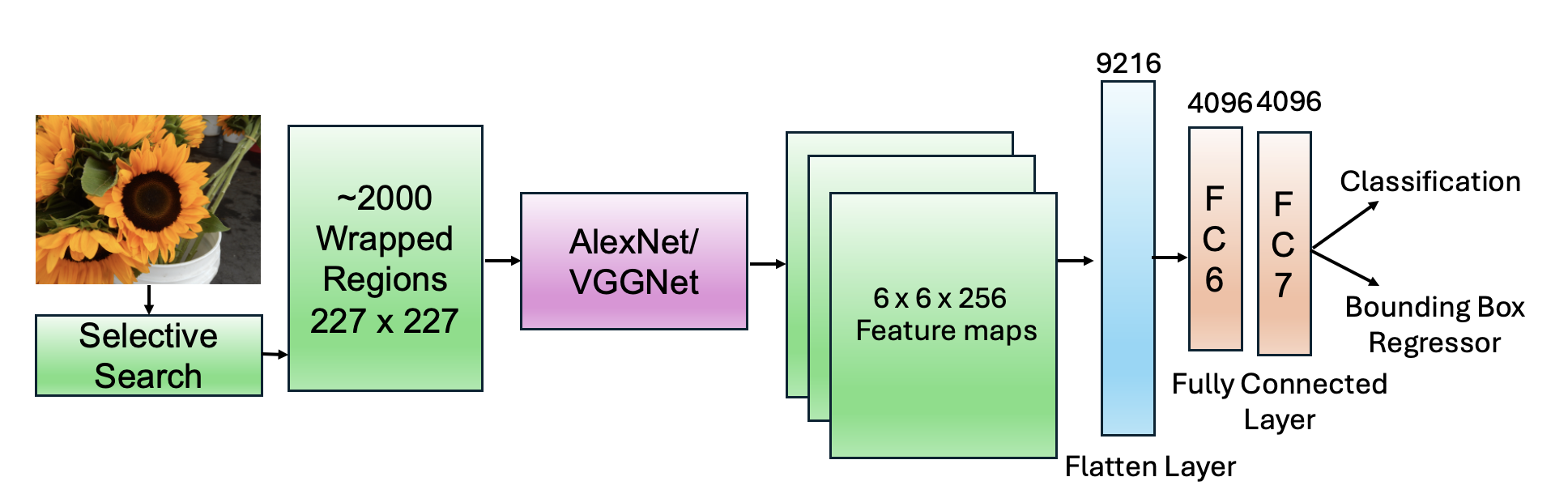}
    \caption{R-CNN Architecture}
    \label{fig:architecture}
\end{figure}

\subsection{Region-based Convolutional Neural Network (R-CNN)}

In 2014, Girshick et al. introduced R-CNN, a two-stage network that combines classical techniques like selective search with CNNs for object detection \cite{girshick2014rich} (see Fig. \ref{fig:architecture}). R-CNN’s training involves three steps: \begin{itemize}
    \item Fine-tune a pre-trained network (e.g., AlexNet) on region proposals generated by selective search.
    \item Train an SVM classifier for object classification.
    \item Use a bounding box regressor to improve localization accuracy.
\end{itemize} 

Selective search generates around 2000 region proposals, each resized to 227x227 pixels for CNN input, reducing the computational cost of exhaustive sliding windows.

Initially, R-CNN achieved 44\% accuracy, improving to 54\% after fine-tuning on warped images. Adding a bounding box regressor boosted accuracy to 58\%, and using VGGNet further increased it to 66\%. While nine times slower than OverFeat, R-CNN’s focus on region proposals reduces false positives, improving accuracy by 10\%.

However, R-CNN has some limitations: 
\begin{itemize}
    \item Feature extraction is performed independently for each proposal, resulting in high computational costs.
    \item The separate stages of proposal generation, feature extraction, and classification prevent end-to-end optimization.
    \item Selective search relies on low-level visual features, struggles with complex scenes, and does not benefit from GPU acceleration.
    \item Despite higher accuracy compared to methods like OverFeat, R-CNN is slower due to these inefficiencies.
\end{itemize}

\subsection{Spatial Pyramid Pooling-Net (SPP-Net)}

\begin{figure}
    \centering
    \includegraphics[width=1\linewidth]{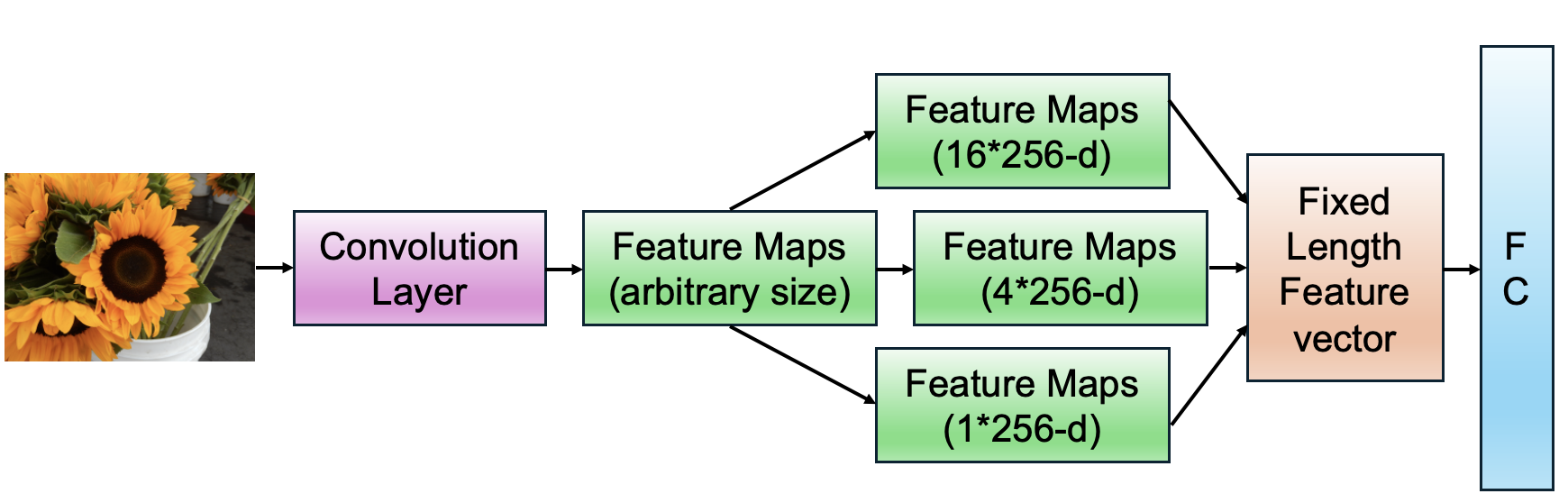}
    \caption{SPP-Net Architecture}
    \label{fig:architecture2}
\end{figure}

In 2015, He et al. introduced SPP-Net to improve detection speed and feature learning over R-CNN \cite{he2015spatial}. Unlike R-CNN, which processes each cropped proposal individually, SPP-Net computes the feature map for the entire image and then applies a Spatial Pyramid Pooling (SPP) layer to extract fixed-length feature vectors (See Fig. \ref{fig:architecture2}). The SPP layer divides the feature map into grids of varying sizes (N $\times$ N), enabling pooling at multiple scales and concatenation of the resulting feature vectors.

SPP-Net allows multi-scale and varied aspect ratio handling without resizing, preserving image details and improving both accuracy and inference speed over R-CNN. However, its multi-stage training hinders end-to-end optimization and requires extra memory for feature storage. Additionally, the SPP layer does not back-propagate to earlier layers, keeping parameters fixed before the SPP layer and limiting deeper learning.

\subsection{Fast Region-based Convolutional Neural Network (Fast R-CNN)}

In 2015, Girshick et al. introduced Fast R-CNN, a two-stage detector designed to improve on SPP-Net’s limitations \cite{girshick2015fast}. Fast R-CNN computes a feature map for the entire image and uses a Region of Interest (ROI) Pooling layer to extract fixed-length features from each region, dividing proposals into a fixed N $\times$ N grid. Unlike SPP, ROI Pooling backpropagates error signals, enabling end-to-end optimization.

After feature extraction, features pass through FC layers, outputting (1) SoftMax probabilities for C+1 classes (including background) and (2) four bounding box regression parameters. Fast R-CNN achieved better accuracy than R-CNN and SPP-Net but still relied on traditional proposal methods.

\subsection{Faster Region-based Convolutional Neural Network (Faster R-CNN)}
\begin{figure}
    \centering
    \includegraphics[width=1\linewidth]{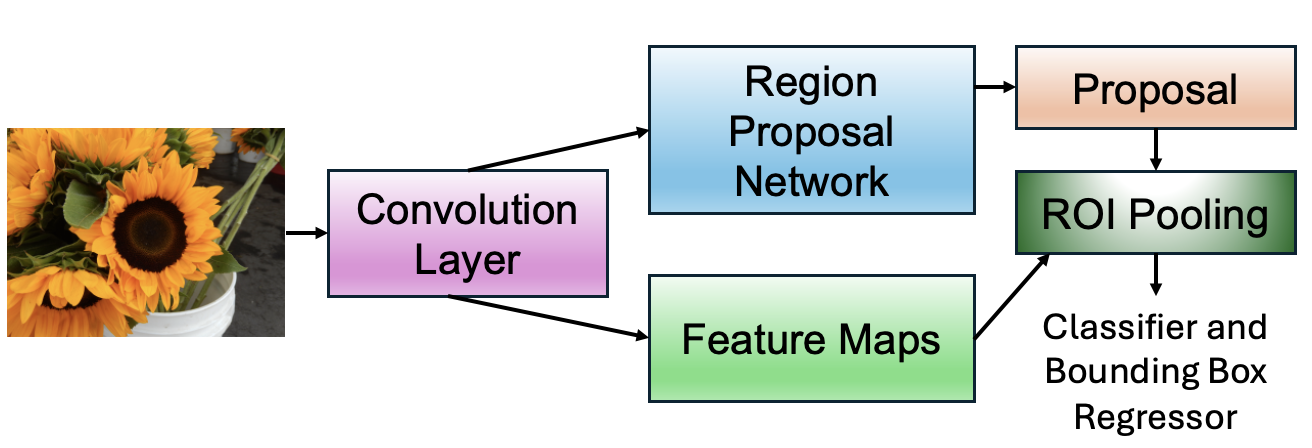}
    \caption{Faster R-CNN Architecture}
    \label{fig:architecture3}
\end{figure}

In 2015, Girshick et al. introduced Faster R-CNN, which utilizes the Region Proposal Network (RPN) to generate object proposals at each feature map position using a sliding window approach (Fig. \ref{fig:architecture3}) \cite{girshick2015fast}. This method shares feature extraction across regions, enhancing efficiency and achieving state-of-the-art results. However, the separate computation for region classification can be inefficient with many proposals, and reliance on a single deep feature map makes detecting objects of varying scales difficult, as deep features are semantically strong but spatially weak, while shallow features are spatially strong but semantically weak.

\subsection{Mask R-CNN}
In 2017, He et al. introduced Mask R-CNN, an extension of Faster R-CNN that performs pixel-level instance segmentation \cite{he2017mask}. It adds a new branch for binary mask prediction to the two-stage pipeline, alongside class and box predictions. This branch uses a fully convolutional network (FCN) atop the CNN feature map. Mask R-CNN also replaces RoIPool with RoIAlign to better preserve spatial accuracy, enhancing mask precision. However, it struggles to detect objects with motion blur in low-resolution images.

\subsection{You Only Look Once (YOLO)}
To increase speed, one-stage models like YOLO (You Only Look Once) were developed, bypassing region proposals. Introduced in 2015 by Redmon et al., YOLO treats detection as a regression task \cite{redmon2016you}. Dividing the image into an S × S grid, YOLO predicts class probabilities, bounding boxes, and confidence scores per cell. This captures context well, reducing false positives, but the grid structure can cause localization errors and struggles with small objects.

YOLO has undergone several iterations, enhancing its performance:
\begin{itemize}
    \item \textbf{YOLOv2/YOLO9000 (2017):} Introduced batch normalization and anchor boxes for improved speed and accuracy \cite{redmon2017yolo9000}.
    \item \textbf{YOLOv3 (2018):} Added multi-scale predictions and residual connections for better detection across various sizes \cite{farhadi2018yolov3}.
    \item \textbf{YOLOv4 (2020):} Enhanced with the CSPDarknet backbone and advanced training techniques, achieving higher precision \cite{bochkovskiy2020yolov4}.
    \item \textbf{YOLOv5 (2021):} Focused on usability, scalability, and deployment flexibility with various model sizes \cite{jocher2021ultralytics}.
    \item \textbf{YOLOv6 (2022):} Optimized for edge devices with improved backbone and attention mechanisms \cite{li2022yolov6}.
    \item \textbf{YOLOv7 (2023):} Employed AutoML techniques for dynamic model optimization, enhancing adaptability \cite{wang2023yolov7}.
    \item \textbf{YOLOv8 (2023):} Incorporated a transformer-based backbone for better detection in dense scenes \cite{yolov8_ultralytics}.
    \item \textbf{YOLOv9 (2024):} Utilized adversarial training to improve robustness against variations \cite{wang2024yolov9}.
    \item \textbf{YOLOv10 (2024):} Implemented real-time feedback loops for dynamic adjustments, boosting accuracy \cite{wang2024yolov10}.
\end{itemize}

These enhancements have established YOLO as a versatile and powerful option for real-time object detection.

\subsection{Single Shot MultiBox Detector (SSD)}
The Single Shot MultiBox Detector (SSD), introduced by Liu et al. in 2016, is a one-stage model that improves on YOLO by using anchors with multiple scales and aspect ratios within each grid cell \cite{liu2016ssd}. Each anchor is refined by regressors and assigned probabilities across categories, with object detection predicted on multiple feature maps for different scales. SSD trains end-to-end with a weighted localization and classification loss, integrating results across maps. Using hard negative mining and extensive data augmentation, SSD matches Faster R-CNN's accuracy while allowing real-time inference.

\begin{table*}[ht]
\centering
\caption{Popular Object Detection Datasets}
\label{table:datasets}
\begin{tabular}{|l|r|r|l|}
\hline
\textbf{Dataset} & \textbf{Number of Images} & \textbf{Number of Classes} & \textbf{Usage} \\ \hline
Pascal VOC & ~0.01 million & 20 & Initial model testing \\ \hline
COCO & ~0.33 million & 80 & Object detection \\ \hline
ImageNet & 1.5 million & 1,000 & Object localization and detection \\ \hline
Open Images & 9.2 million & 600 & Object localization \\ \hline
\end{tabular}
\end{table*}

\begin{table*}[ht]
\centering
\caption{Quantitative Performance Comparison of Object Detection Models on different Dataset}
\label{table:model_performance}
\begin{tabular}{|l|c|c|c|c|c|c|}
\hline
\textbf{Model} & \textbf{Pascal VOC (mAP)} & \textbf{COCO (mAP)} & \textbf{ImageNet (mAP)} & \textbf{Open Images (mAP)} & \textbf{Inference Speed (FPS)} & \textbf{Model Size (MB)} \\ \hline
RCNN & 66\% & 54\% & 60\% & 55\% & $\sim$5 FPS & 200 \\ \hline
Fast RCNN & 70\% & 59\% & 63\% & 58\% & $\sim$7 FPS & 150 \\ \hline
Faster RCNN & 75\% & 65\% & 68\% & 63\% & $\sim$10 FPS & 180 \\ \hline
Mask RCNN & 76\% & 66\% & 69\% & 64\% & $\sim$8 FPS & 230 \\ \hline
YOLO & 72.5\% & 58.5\% & 61.5\% & 57.5\% & $\sim$45--60 FPS & 145 \\ \hline
SSD & 75\% & 63.5\% & 66.5\% & 61.5\% & $\sim$19--46 FPS & 145 \\ \hline
\end{tabular}
\end{table*}

\section{APPLICATIONS}

Object detection, powered by CNN, has diverse applications, spanning from targeted advertising to self-driving cars and beyond. It is utilized for handwritten digit recognition, Optical Character Recognition (OCR), face detection, medical image analysis, sports analytics, and more.

\begin{itemize}
    \item \textbf{Optical Character Recognition (OCR):}
OCR converts images of text into machine-encoded text, facilitating tasks such as document digitization, automated data entry, and cognitive computing.

\item \textbf{Self-Driving Cars:}
Object detection is essential for autonomous vehicles to detect and classify objects such as cars, pedestrians, traffic lights, and road signs.

\item \textbf{Object Tracking:}
Used in tracking objects in videos, object detection has applications in surveillance, traffic monitoring, and sports analytics.

\item \textbf{Face Detection and Recognition:}
Widely employed in computer vision, object detection is used for social media image tagging and biometric security systems.

\item \textbf{Object Extraction from Images or Videos:}
Facilitates segmentation and meaningful representation of images, potentially enabling applications like video object extraction.

\item \textbf{Digital Watermarking:}
Embed markers into digital signals for copyright protection and authentication purposes.

\item \textbf{Medical Imaging:}
Assists clinicians in diagnosis and therapy planning, particularly in tracking anatomical objects.

Object detection technology continues to evolve, promising further advancements and expanding its applications across various industries.
\end{itemize}

\section{POPULAR DATASET}
Key datasets in object detection include Pascal VOC \cite{everingham2010pascal}, COCO \cite{lin2014microsoft}, ImageNet \cite{deng2009imagenet}, and Open Images \cite{kuznetsova2020open}. Pascal VOC (Visual Object Classes) offers a manageable size, balancing complexity and computational efficiency, making it ideal for testing. COCO (Common Objects in Context) provides extensive annotations with multiple objects per image, including segmentation and key points. ImageNet, primarily used for classification, also includes object detection annotations. Open Images, with over 600 labeled categories, stands out for its large scale, offering both bounding box annotations and segmentation masks. Table~\ref{table:datasets} summarizes the key attributes of each dataset, emphasizing their unique features and primary usage. Table~\ref{table:model_performance} provides a comparison of the performance of RCNN, Fast RCNN, Faster RCNN, Mask RCNN, YOLO, and SSD on these datasets in terms of mAP, inference speed (measured in Frames Per Second, or FPS), and model size.

\section{Evaluation Metrics}
Object detection models are assessed using several key metrics: Intersection over Union (IoU), Mean Average Precision (mAP), Precision, Recall, Confidence Score (CS), F1 Score, and Non-Maximum Suppression (NMS). Table~\ref{table:evaluation_metrics1} summarizes these metrics, highlighting their limitations and potential biases.

\subsection{Intersection over Union (IoU)}
IoU measures the overlap between the predicted and ground truth bounding boxes, calculated as the ratio of the intersection area to the union area:

\[
\text{IoU} = \frac{\text{Area of Intersection}}{\text{Area of Union}}
\]

\subsection{Mean Average Precision (mAP)}
mAP evaluates model performance by averaging the precision across all classes. The Average Precision (AP) is computed as:

\[
\text{AP} = \frac{\sum_{k=1}^n \left( P(k) \times \text{Precision at Recall}(k) \right)}{n}
\]

where \( P(k) \) is the change in recall from the previous highest recall, and precision at recall \( k \) is the maximum precision observed at any recall level \( j \) where \( j \geq k \).

\subsection{Precision and Recall}
\textit{Precision} is the ratio of true positives to all positive predictions, while \textit{Recall} is the ratio of true positives to all ground truth positives.

\subsection{Confidence Score (CS)}
The Confidence Score reflects the model's certainty that a predicted bounding box contains the correct object. Higher scores indicate greater accuracy and help set thresholds for accepting or rejecting detections.

\subsection{Non-Maximum Suppression (NMS)}
Non-Maximum Suppression refines bounding box predictions by sorting them by confidence scores and selecting the highest one while suppressing overlapping boxes. This process ensures each object is detected once, improving accuracy and efficiency.

\begin{table*}[htbp]
\centering
\caption{Evaluation Metrics: Limitations and Potential Biases of Object Detection Models}
\label{table:evaluation_metrics1}
\begin{tabular}{|l|l|p{5cm}|p{5cm}|}
\hline
\textbf{Model} & \textbf{Metrics Used} & \textbf{Limitations} & \textbf{Potential Biases} \\ \hline
RCNN & IoU, mAP, Precision, Recall, F1 Score & 
- Separate region proposal step slows inference. 
- High memory usage due to multiple stages. 
& 
- Favors larger objects due to reliance on selective search. 
- Struggles with scale variations and densely packed objects. \\ \hline

Fast RCNN & IoU, mAP, Precision, Recall, F1 Score & 
- Dependent on external region proposals. 
- Not optimized for real-time applications. 
& 
- Similar biases as RCNN: prefers larger and well-separated objects. 
- Performance drops in high-density scenes. \\ \hline

Faster RCNN & IoU, mAP, Precision, Recall, F1 Score & 
- More complex architecture with integrated Region Proposal Network (RPN). 
- Requires careful hyperparameter tuning. 
& 
- Favors objects with distinct features detectable by RPN. 
- Limited accuracy on small or thin objects compared to single-shot models. \\ \hline

Mask RCNN & IoU, mAP, Precision, Recall, F1 Score & 
- Increased computational overhead from mask prediction. 
- Longer training times. 
& 
- Bias towards classes with abundant and detailed segmentation data. 
- Misses small or occluded objects in segmentation masks. \\ \hline

YOLO & IoU, mAP, Precision, Recall, Confidence Score & 
- Lower detection accuracy on small objects. 
- Struggles with overlapping objects and crowded scenes. 
& 
- Prioritizes objects at the center of the image. 
- Predefined grid may miss objects at image edges. \\ \hline

SSD & IoU, mAP, Precision, Recall, Confidence Score & 
- Performance degrades on very small objects. 
- Limited by predefined anchor box scales and aspect ratios. 
& 
- Bias towards predefined anchor boxes, affecting generalization for unseen scales. 
- Struggles with variable object shapes and sizes not covered by anchor boxes. \\ \hline
\end{tabular}
\end{table*}

\section{Discussion and Future Directions}

This review examined prominent object detection models, classifying them into classical computer vision techniques and CNN-based methods. While recent CNN architectures have significantly improved accuracy to below 5\%, they also increase complexity and resource demands. Traditional models like Deformable Part Models (DPMs) are shallower and more lightweight, making them better suited for edge deployment compared to modern deep learning architectures like AlexNet and VGGNet.

Key future directions for object detection include:
\begin{itemize}
    \item Speed-Accuracy Trade-off: Enhancing both accuracy and speed for real-time, low-power applications.
    \item Tiny Object Detection: Improving the detection of small objects in areas such as wildlife monitoring and medical imaging.
    \item 3D Object Detection: Leveraging 3D sensors for applications in augmented reality and robotics.
    \item Multi-modal Detection: Integrating visual and textual sources for better accuracy in complex scenarios.
    \item Few-shot Learning: Developing models that can effectively detect objects from limited examples, particularly in low-resource settings.
\end{itemize}

This review aims to foster interest in advancing object detection models and to inspire innovation to address current limitations, including minimizing environmental impacts.
\section*{Acknowledgment}
This study was partly supported by the West Chester University faculty development fund.

\bibliographystyle{IEEEtran}
\bibliography{main.bib}

\end{document}